\documentclass[journal]{IEEEtran}

\usepackage{amsmath,amsfonts}
\usepackage{array}
\usepackage{textcomp}
\usepackage{stfloats}
\usepackage{url}
\usepackage{verbatim}
\usepackage{graphicx}
\def\BibTeX{{\rm B\kern-.05em{\sc i\kern-.025em b}\kern-.08em
    T\kern-.1667em\lower.7ex\hbox{E}\kern-.125emX}}
\usepackage{balance}
\IEEEoverridecommandlockouts
\usepackage{blindtext}
\usepackage{amssymb}
\usepackage{amsmath}
\usepackage{cite}
\usepackage{citesort}
\usepackage{multicol}
\usepackage{amsfonts}
\usepackage{color}
\usepackage{subfigure}
\usepackage{epstopdf}
\DeclareGraphicsExtensions{.jpg,.png}
\usepackage{amssymb}
\usepackage{comment}
\usepackage{amsmath}
\allowdisplaybreaks[4]
\usepackage{booktabs}
\usepackage{multirow}
\usepackage{bmpsize}
\usepackage{algorithm}
\usepackage{algorithmicx}

\usepackage{lipsum}
\usepackage{stfloats}
\usepackage{algpseudocode}
\usepackage{amsfonts}

\newtheorem{theorem}{\textbf{Theorem}}

\newtheorem{lemma}{\textbf{Lemma}}

\newtheorem{corollary}{\textbf{Corollary}}

\makeatletter
\def\ScaleIfNeeded{%
\ifdim\Gin@nat@width>\linewidth \linewidth \else \Gin@nat@width
\fi } \makeatother

\begin{document}
%

\title{{Scalable Dexterous Robot Learning with AR-based Remote Human-Robot Interactions}}

\author{Yicheng Yang, Ruijiao Li, Lifeng Wang, Shuai Zheng, Shunzheng Ma, Keyu Zhang,\\Tuoyu Sun,  Chenyun Dai, Jie Ding, and Zhuo Zou
}

\maketitle

\begin{abstract}
This paper focuses on the scalable robot learning for manipulation in the dexterous robot arm-hand systems, where the remote human-robot interactions via augmented reality (AR) are established to collect the expert demonstration data for improving efficiency. {In such a system, we present a novel method to address the general manipulation task problem.} Specifically, the proposed method consists of two phases: i) In the first phase for pretraining, the policy is created in a behavior cloning (BC) manner, through leveraging the learning data from our AR-based remote human-robot interaction system; ii) In the second phase, a contrastive learning empowered reinforcement learning (RL) method is developed to obtain  more efficient and robust policy than the BC, and thus a projection head is designed to accelerate the learning progress. An event-driven augmented reward is adopted for enhancing the safety. To validate the proposed method, both the physics simulations via PyBullet and real-world experiments are carried out. The results demonstrate that compared to the baselines, {our method not only significantly speeds up the training process,} but also achieves much better performance in terms of the success rate for fulfilling the manipulation tasks. By conducting the ablation study, it is confirmed that the proposed RL with contrastive learning overcomes policy collapse. Supplementary demonstrations are available at https://cyberyyc.github.io/.
\end{abstract}

\begin{IEEEkeywords}
Augmented reality, contrastive learning, human-robot interactions, imitation learning, reinforcement learning.
\end{IEEEkeywords}

\section{Introduction}
Artificial intelligence (AI) is essential for tackling the complicated robotics problems across the manipulation tasks, environments and embodiments, and the AI-driven robots become more  capable of performing a variety of tasks in daily lives~\cite{Billard2025}. With the rapid development of AI and robotics, robots' capabilities and scalability heavily rely on the embodied AI arisen from human-robot interactions with their surroundings~\cite{Jiafei2022}. In particular, the deployment of the dexterous robotic arm-hand system requires sophisticated control designs since they have high-dimensional degree-of-freedoms (DoFs)~\cite{Handa2020}. The machine learning algorithms have attracted significant attention, which are implemented to acquire efficient robotic manipulation policies~\cite{Kroemer2021}. Therefore, many research contributions have leveraged the machine learning to enhance the dexterity in the robotic systems~\cite{zhaochao2025,GaoXiao2024}.

Imitation learning aims to make robotic manipulation through mimicking the expert demonstrations. Such an approach can quickly learn a complex interaction trajectory and have been applied in the autonomous systems~\cite{Tianhao2018,Edward2021}.  To improve the autonomous driving performance, a modified dataset aggregation based imitation learning is proposed in~\cite{MichaelKelly}. The work of \cite{Jianlan2024} investigates the cable routing task via imitation learning, in which a hierarchical cable routing policy is designed. In the dexterous robot manipulation task for deformable objects, \cite{Heecheol2024} presents a goal-conditioned dual-action deep imitation learning to acquire the expert skills. However, the imitation learning methods such as behavior cloning could be degraded by the data mismatch and compounding errors~\cite{MichaelKelly}. To address these issues, \cite{TZZhao2023} adopts the action chunking with transformers to learn policy from expert demonstrations. When the robots recognize the scenes via behavior cloning,  \cite{Shaopeng2025} develops a revision and prediction method to reduce the risks of the compounding error. In addition, the combination of imitation learning with reinforcement learning (RL) enables fast learning and lower overhead for manipulation with dexterous hands~\cite{Rajeswaran2018,QLiu2023}.

Since the generalizability of imitation learning requires high-quality expert demonstrations through the human-robot interactions, many efforts have been devoted to the teleoperated robotic designs~\cite{Kourosh2023,Yang_Chao}. There are many teleoperation modalities such as the visual and haptic feedback. A virtual reality (VR) based teleoperated robotic system is built in~\cite{Tianhao2018}, where human demonstrations are made in the virtual environment. With the help of VR demonstrations, \cite{Giuseppe25} proposes an incremental learning method for grasping and lifting tasks. In~\cite{Walker2019}, augmented reality (AR) is applied to create the virtual robot surrogates, and such a design helps participants teleoperate a robot more effectively.  To reduce the cost of high degree-of-actuation teleoperation, a pure vision-based real-time hand tracking method is adopted to translate human hand motion to robot motion in~\cite{Handa2020}. The work of \cite{ding2024bunnyvision} leverages a VisionPro headset to  teleoperate high-DoF  hand-arm systems and improves the imitation learning performance. By applying AR in the human-robot collaboration, \cite{Junling2024} finds that errors in task completion could be minimized. The teleoperation system in \cite{Mukashev2025} employs the AR immersion and tactile sensors, where AR generates the virtual interaction with the objects. In 6G, extended reality (XR) consisting of VR and AR is one of the most transformative services~\cite{huming}, and it turns out that XR based teleoperation is expected to further facilitate the human-robot interaction and robotic manipulation.

While the teleoperation assisted imitation learning can improve the robot cognitive skills, collecting a large amount of high-quality expert demonstrations may be very time-consuming and imperfect demonstrations deteriorate robotic manipulation tasks~\cite{Maryam2024}. To this end, deep RL is viewed as a scalable approach to handling the dexterous manipulation~\cite{ZhuHenry}, thus  many RL architectures and their variants are studied in the literature. In~\cite{Deirdre2018}, the deep Q-learning and deep deterministic policy gradient (DDPG) algorithms are employed for grasping task with the gripper, and the simulations are operated in the Bullet simulator. Through introducing the continuous-action generalization of Q-learning, \cite{Kalashnikov2018} proposes an off-policy training method to improve the success rate of grasping unseen objects. In the human-robot collaboration,  \cite{Shamouty2020} shows that DDPG policies with hybrid reward architecture could reduce the hazards incurred by the robots   and performance evaluations are presented based on the MuJoCo physics simulator. Joint  design of graph convolutional networks and recurrent Q-learning is exploited to complete a collaborative packaging task in \cite{Ghadirzadeh}, where the effects of the perception uncertainties are analyzed. In \cite{Guofei2021},  expert demonstrations are employed to prompt the augmented task-oriented guiding reward for robotic locomotion, however, it is assumed that the expert demonstrations are obtained from the trained RL agent. {The work of \cite{Dmytro2025} provides a proximal policy optimization (PPO) approach for dexterous grasp manipulation with an anthropomorphic hand.} In addition, recent study \cite{Sharifi} provides a comprehensive overview of RL methods for assistive and rehabilitation robotic manipulation. The fusion of RL with imitation learning can significantly improve the sample efficiency~\cite{Rajeswaran2018}, e.g., \cite{Albert2022} shows that by using a small number of expert demonstrations for pretraining the RL agent, the robot learning efficiency from pixels can be enhanced; \cite{seo2022masked} points out that solving more challenging visual robotic manipulation task problems requires both imitation learning and RL algorithms. In \cite{Hansen2023}, a model-based RL with behavior cloning is applied to address the complex visuo-motor control problem. An imitation bootstrapped RL is presented in \cite{hengyuan2024}, and its core idea is that the twin delayed DDPG (TD3) can gain the benefits from bootstraping the target values based on the imitation policy and RL policy.  In the vision-based agile quadrotor flight scenario, \cite{xing2024bootstrapping}  demonstrates that the imitation bootstrapped RL enables successful navigation using solely RGB images.

Prior work finds that RL is unable to learn effective representations from high-dimensional observations and additional representation learning can improve the RL policy~\cite{eysenbach2022contrastive}. Contrastive learning is a promising approach to learning good representations~\cite{oord2019contrastive,chenting2020}, which can be employed to construct better RL algorithms. Early work~\cite{MichaelLaskin20} integrates contrastive learning with model-free RL  for pixel-based control tasks, and highlights  that its method achieves state-of-the-art data-efficiency. Since consecutive video frames are correlated, \cite{ZhuJinhua} extends the work of~\cite{MichaelLaskin20} to train the pixel-based RL policy with masked contrastive learning model, where masked pre-training and transformer module are applied to leverage the correlations among video frames for learning better representations. In~\cite{liu2021return-based}, a new contrastive loss function is introduced to learn the return-related representations for speeding up the RL algorithm.  By  contrasting different-age trajectory experiences, \cite{WangDi2023} trains the contrastive RL policy for balancing the new and the old experience in the PyBullet robotics environments. In the online RL, \cite{ShuangQiu2022} adopts the contrastive learning to constrain the state transition. In \cite{ChengLu2023}, a contrastive energy prediction method is proposed to learn the desired guidance for the sampling procedure in the offline diffusion-based RL, and achieves the similar performance compared to the classifier guidance. Recent work \cite{shan2025contradiff} aims to guide the diffusion-based RL policy toward high-returns via a contrast mechanism. To meet the stable self-supervised RL, \cite{Chongyi2024} leverages the contrastive learning to evaluate the similarity between the current state  representation and future state  representation for retaining task-relevant information. The on-robot contrastive RL proposed by \cite{biza25onrobot} provides a goal-contrastive reward for improving the reward model and  sample efficiency. Therefore, existing studies have indicated that contrastive learning is an important means to improve the RL performance. { To the best of our knowledge, the integration of contrastive RL with imitation learning for dexterous robotic manipulation has not been conducted yet.}

In this work, we present a novel method for manipulation tasks in the dexterous robot arm-hand systems. To tackle the general manipulation task problem, an efficient imitation and contrastive learning assisted  RL algorithm is proposed. Therefore, our main contributions are three-fold:
\begin{itemize}
  \item \textbf{AR-based teleoperated robotic system for learning data collection:}  To speed up the RL training process and guide exploration, a behavior cloning based pretraining is adopted by solving a simple regression problem. Therefore, we create a general AR-based teleoperated dexterous robotic system for collecting expert demonstration data. In such a teleoperation system, we employ the Unity platform at the edge  server  to make sure  that different types of AR headsets and robotic systems are compatible and they can be wirelessly and remotely connected via Unity. The AR headset with its camera captures the expert behavior, which are replicated by the dexterous robot via teleoperation, thus the learning data are collected.
  \item \textbf{Imitation and contrastive learning assisted RL policy:} { While existing works~\cite{Rajeswaran2018,QLiu2023,seo2022masked,Hansen2023,hengyuan2024} have adopted imitation learning for pretraining the RL policy through using supervised learning, the benefits of policy pretraining could be erased and policy collapse occurs during the RL training~\cite{Albert2022,Hansen2023}. To address these challenging issues, we develop a novel contrastive learning approach to guiding the model-free RL policy, and  introduce a projection head to enforce an expert-preferred constraint on the RL policy, which has not been reported in the literature.}  As such, the agent's state-actions are constrained towards high-return expert state-actions. { We  design the event-driven and hand pose adjustment regularizers for safety enhancement and efficient target grasp representation.} In addition, only a small number of human demonstrations (15 expert trajectories applied in the experiments) are collected  and the RL training is operated in the PyBullet simulation environment. Compared to the PPO~\cite{PPO2017} and soft Actor-Critic (SAC)~\cite{haarnoja18b}  algorithms, the proposed algorithm consumes much lower training time and overcomes the policy collapse issue.
  \item \textbf{Experiments in the simulated environment and the real-world:} To validate the efficacy of proposed algorithm, we make the experiments in the simulator and real-world. Both the simulation results and real-world trials demonstrate that our proposed algorithm significantly improves the RL policy and achieves higher success rate for completing dexterous manipulation tasks  than the baselines. In the ablation study, it is confirmed that behavior cloning based policy pretraining  plays a dominant role in reducing the RL training time and contrastive learning can further improve the model-free RL performance.
\end{itemize}

The rest of the paper is organized as follows: Section II describes the general dexterous robot arm-hand system model. Section III proposes an imitation and contrastive learning assisted RL algorithm.  Section IV  presents numerical results and the concluding remarks are given in Section V.

\section{System Descriptions}\label{System_description}
In this section, a general manipulation task problem is studied in the dexterous robot arm-hand systems. Moreover, a remote human-robot interaction system via AR-based visual feedback is established to collect the learning data, so as to support the data-driven solutions of the manipulation task problems.
\subsection{Overall Design}
The dynamics of an arbitrary dexterous robot arm-hand system is given by
\begin{align}\label{eq1}
{\bf{\dot s}} = f\left( {{\bf{s}},\mathbf{a}} \right),
\end{align}
where $\bf{s}$ is the system's state vector including the joint angles of the robot's arm and the end-effector pose; $f$ is the non-linear function; $\mathbf{a}$ is the control input consisting of the joint angle biases and hand pose for the robot's arm and hand. The robot seeks a motion to reach the target object for manipulation tasks with its dexterous arm-hand, and the manipulation task problem is formulated as
{ \begin{align}\label{prob1}
 &\mathop {\min }\limits_{\mathbf{a}} \sum\limits_t {\left\| {\mathbf{s}_t  - \mathbf{s}_\mathcal{T} } \right\|^2 } +  \sum\limits_i \xi_i g_i\left({\bf{s}},\mathbf{a}\right)  \nonumber \\
&{\rm s.t.} \quad {\bf{\dot s}} = f\left( {{\bf{s}},\mathbf{a}} \right), \mathbf{s}_t \in \mathcal{S}, \forall t,
\end{align} }
where $\mathbf{s}_t$ is the state at the time step $t$; $\mathbf{s}_\mathcal{T}$ is the target joint and pose positions; $\mathcal{S}$ denotes the feasible state space with safety concerns; the  regularizer $g_i\left({\bf{s}},\mathbf{a}\right)$ with the weight $\xi_i$ aims to improve the system's robustness. In general, problem (2) is non-convex and the regularizers may be various across different target objects and environments. To  improve the sample efficiency and reduce the training time for robot control policies, behavior cloning is adopted to learn policies from expert demonstrations in a supervised learning manner, which can be integrated into the regularization. Therefore, we first introduce a general AR-based learning data collection system.

\begin{figure}[htbp]
\centering
    \subfigure[The hardware and software components in our teleoperated dexterous robotic system for remote human-robot interactions.]{
         \centering
         \includegraphics[width=3.4in,height=2.8in]{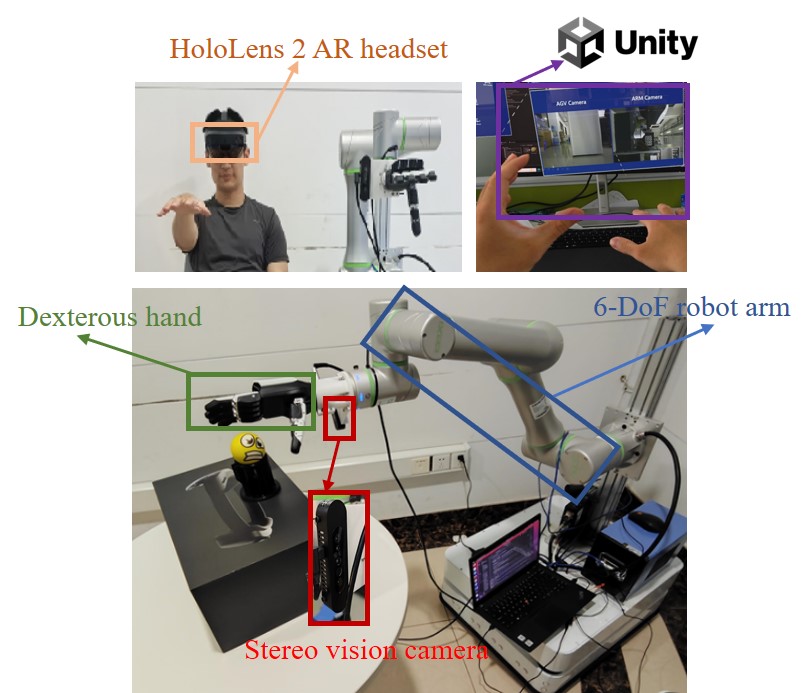}
      \label{fig1a}
     }
         \subfigure[The dexterous robot replicates the expert behavior.]{
         \centering
         \includegraphics[width=2.4 in,height=1.9 in]{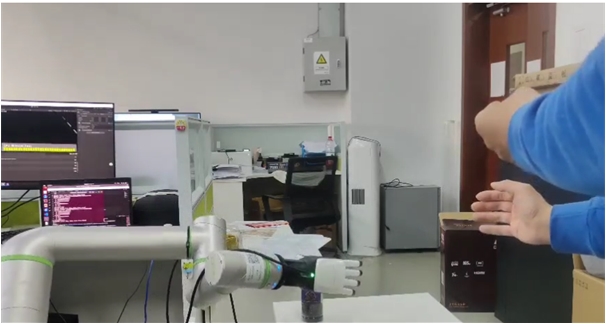}
      \label{fig1b}
    }
    \caption{Illustration of a learning data collection system via AR: The expert wears the AR headset to teleoperate the dexterous robot for manipulation tasks, and the expert demonstration data are collected for model training.}
 \label{fig11}
\end{figure}
\subsection{AR-based Remote Human-robot Interactions}

To collect the learning data from expert demonstrations, we create a dexterous robot arm-hand teleoperation system via AR, and utilize the HoloLens 2 AR headset to detect and track the expert's hand pose, as shown in Fig. 1. Both the AR headset and dexterous robot are connected to the Unity platform{\footnote{  The Unity-based application leverages the Mixed Reality Toolkit (MRTK) to enable interaction with the HoloLens 2 AR headset and establishes communication with the robot system through the ROS TCP Connector.}} via WiFi or 5G for remote human-robot interactions, and the robot replicates the expert actions for manipulation tasks, namely the expert behavior is  mapped to the physical robot's kinematics. A stereo vision camera is mounted on the robot hand, in order to detect the target object's positions. As such, our remote teleoperation system is able to capture diverse expert demonstrations, and is suitable for a wide range of robot manipulation tasks. After obtaining the expert demonstration data, {a lightweight multilayer perceptron (MLP) is leveraged to solve the regression problem such that the robot control policies are aligned with expert preferences.}

Since the non-convex manipulation task problem \eqref{prob1} is generic and its global optimum is challenging to obtain, we develop an RL approach to solving it. To substantially speed up the training process and enhance the sample efficiency, both the behavior cloning and contrastive learning are employed to improve the RL performance.
\begin{figure*}
\centering
 \includegraphics[width=6.0in,height=2.1in]{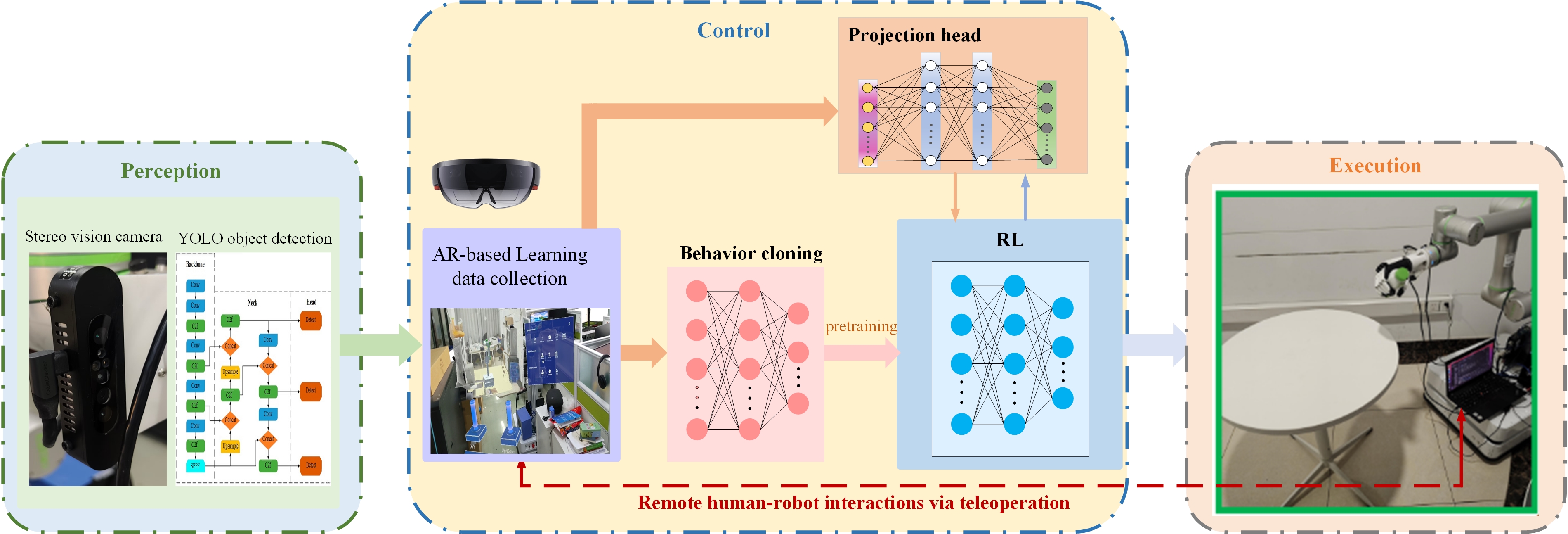}
\caption{Design structure of the proposed algorithm: Differing from the existing algorithms, the expert demonstration data are leveraged for behavior cloning based pretraining and augmented RL policy with the help of projection head.}
 \label{fig2}
\end{figure*}
\section{Algorithm Design}\label{sec:algorithm}

In the dexterous robot arm-hand systems, we develop a general algorithm design for fulfilling a variety of the manipulation tasks. As shown in Fig. 2, the proposed algorithm with visual perception undergoes two phases for learning-based manipulation control, i.e., in the first phase, the initial policy is obtained by solving the regression problem in a behavior cloning manner, namely the expert demonstrations are learned for pretraining; in the second phase, a contrastive learning enhanced RL approach is developed to address the whole constrained problem (2). Since behavior cloning has been well studied~\cite{TZZhao2023}, we seek the RL policy for tackling the problem (2) in the second phase.

Problem (2) is transformed into the RL problem, which maximizes the expected sum of the entropy augmented rewards, i.e.,
\begin{align}\label{eq3}
\mathop {\max }\limits_{\mathbf{a}} \mathcal{J}=
\sum\limits_{t = 0}^T {\mathbb{E}_{\left( {\mathbf{s}_t ,\textbf{a}_t } \right) \sim \rho _\pi  } \Big\{ {\gamma ^t \left[ {r\left( {\mathbf{s}_t ,\mathbf{a}_t } \right) + \alpha H\left( {\pi \left( {\left. \cdot \right|\mathbf{s}_t } \right)} \right)} \right]} \Big\}},
\end{align}
where $T$ is the length of time-steps; the agent's total reward is $r\left( {\mathbf{s}_t ,\mathbf{a}_t } \right)$ with the state $\mathbf{s}_t$ and its action $\mathbf{a}_t\sim \pi \left( {\left. \cdot \right|\mathbf{s}_t } \right)$ from interacting with the environment;  $\rho _\pi $ is the state distribution over the policy $\pi\left(\left. \mathbf{a} \right|\mathbf{s}\right)$; $\gamma $ is the discount factor; $\alpha$ is the  temperature parameter, and $H\left( {\pi \left( {\left. \cdot \right|\mathbf{s}_t } \right)} \right)$ is the entropy function. In light of the objective and constraints in (2), the total reward $r\left( {\mathbf{s}_t ,\mathbf{a}_t } \right)$ is calculated as
{\begin{align}\label{eq4}
r\left( {\mathbf{s}_t ,\mathbf{a}_t } \right)=&-\left\| {\mathbf{s}_t  - \mathbf{s}_\mathcal{T} } \right\|^2+ \xi_1 r_{\rm smooth} +\xi_2  r_{\rm succ}+ \xi_3  r_{\rm pose},
\end{align}
where $r_{\rm smooth}= \left\| {\mathbf{s}_{t-1}  - \mathbf{s}_\mathcal{T} } \right\|^2-\left\| {\mathbf{s}_t  - \mathbf{s}_\mathcal{T} } \right\|^2$ is the motion smoothness reward  with the weight $\xi_1$; $r_{\rm succ}$ is the event-driven reward with the weight $\xi_2$ for safety enhancement; $r_{\rm pose}$ is the hand pose adjustment reward  with the weight $\xi_3$. The event-driven reward  $r_{\rm succ}$ is a piecewise function to evaluate the effects of events including the success of completing a manipulation task, collision and hand-object contact without mission completion, and it is given by
 \begin{align}\label{eq5}
 r_{\rm succ}=\left\{ \begin{array}{l}
 {\mathcal{Z}}_1 ,\,\quad~~\mathrm{success}, \\
 {-\mathcal{Z}}_2 ,\quad \mathrm{collision},\\
 {-\mathcal{Z}}_3 ,\quad \rm{hand\texttt{-}object~contact}, \\
 0,\quad~~~~\mathrm{otherwise},\\
 \end{array} \right.
\end{align} }
where $\mathcal{Z}_1$, $\mathcal{Z}_2$ and $\mathcal{Z}_3$ are the predefined positive values. The hand pose adjustment award $r_{\rm pose}$ is defined as
 \begin{align}\label{eq6}
 r_{\rm pose}=\left\{ \begin{array}{l}
 {\mathcal{Z}}_4 \left(\cos \psi _t - \Lambda_{\rm th}\right),\,\quad\mathrm{if~
\cos \psi _t \leq \Lambda_{\rm th}}, \\
 {\mathcal{Z}}_5 \left(\cos \psi _t-\Lambda_{\rm th}\right),\quad \mathrm{otherwise},\\
 \end{array} \right.
\end{align}
 where $\mathcal{Z}_4$ and $\mathcal{Z}_5$ are the predefined positive values; $\Lambda_{\rm th}$ is the geometric reasoning parameter, and the geometric alignment between the hand and object is evaluated by
 \begin{align}\label{eq7}
\cos \psi _t  = \frac{{\left\langle {\bf{n }}_{\rm hand} ,\left({{\boldsymbol{\nu }}_{\rm object} } -{\boldsymbol{\nu }}_{\rm hand}\right) \right\rangle }}{{\left\| {{\bf{n }}_{\rm hand} } \right\|\left\| {{\boldsymbol{\nu }}_{\rm object} } -{\boldsymbol{\nu }}_{\rm hand}\right\|}},
\end{align}
where ${\bf{n }}_{\rm hand}$ is the dexterous hand's normal vector; ${\boldsymbol{\nu }}_{\rm hand}$ and ${\boldsymbol{\nu }}_{\rm object}$ represent the position centers of the dexterous hand and the target object, respectively; the inner product operator is denoted as $\left\langle \cdot , \cdot \right\rangle$.

\begin{figure*}
\centering
 \includegraphics[width=6.0in,height=2.2in]{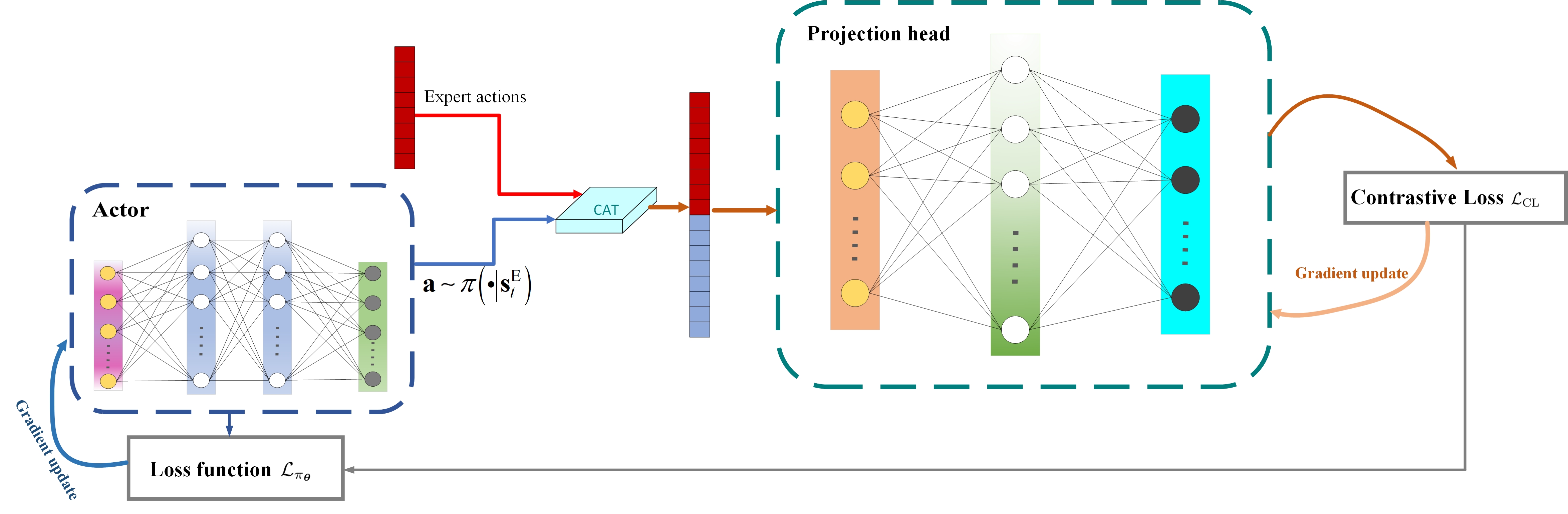}
\caption{The proposed projection head for improving the sample-efficiency and preventing the policy collapse in the actor network.}
 \label{fig3}
\end{figure*}
 Inspired by the SAC policy with high sample efficiency and stability~\cite{haarnoja18b},  the RL problem \eqref{eq3} is addressed by jointly optimizing the soft Q-function  and policy. We employ two Q-functions $Q_{\boldsymbol{\phi}_i}$ with the parameters $\boldsymbol{\phi}_i$ ($i\in\{1,2\}$) for the critic network, and optimize them through minimizing the Bellman residual, which is given by
\begin{align}\label{eq8}
\mathop {\min }\limits_{\boldsymbol{\phi}_i} \mathcal{L}_{\boldsymbol{\phi}_i}=\mathbb{E}_{{\left(\mathbf{s}_t ,\mathbf{a}_t \right)}\sim \mathcal{D}} \Big\{\frac{1}{2}\left(Q_{\boldsymbol{\phi}_i}\left( {\mathbf{s}_t ,\mathbf{a}_t } \right)-\overline{Q}\left( {\mathbf{s}_t ,\mathbf{a}_t } \right)\right)^2 \Big\},
\end{align}
where $\mathcal{D}$ is the replay buffer, and  the target Q-function $\overline{Q}$ is
\begin{align}\label{eq9}
\hspace{-0.3cm}\overline{Q}\left( {\mathbf{s}_t ,\mathbf{a}_t } \right)=&r\left( {\mathbf{s}_t ,\mathbf{a}_t } \right)+\gamma \mathbb{E}_{\left( {\mathbf{s}_{t+1},\textbf{a}_{t+1} } \right) \sim \rho _\pi  } \Big\{\mathop {\min }\limits_{i\in\{1,2\}}Q_{\boldsymbol{\bar{\phi}}_i} \left( {\mathbf{s}_{t+1},\textbf{a}_{t+1} } \right)\nonumber \\
&\quad\quad\quad~~~~~~~~-\alpha\log\left(\pi\left(\left. \textbf{a}_{t+1} \right|\mathbf{s}_{t+1}\right)\right)\Big\},
\end{align}
where the Q-function $Q_{\boldsymbol{\bar{\phi}}_i}$ has the parameters ${\boldsymbol{\bar{\phi}}_i}$ (${\boldsymbol{\bar{\phi}}_i}$ is the exponentially moving average of the parameters $\boldsymbol{\phi}_i$).

In the original policy improvement step, the policy is learned by minimizing the expected
Kullback-Leibler divergence $\mathrm{D}_{\rm KL}$, which is expressed as
\begin{align}
\mathop {\min }\limits_{\pi} \mathbb{E}_{\mathbf{s}_t \sim \mathcal{D} } \Bigg\{ \mathrm{D}_{\rm KL}\left( \pi\left(\left. \cdot \right|\mathbf{s}_t\right) \parallel \exp\left(\frac{1}{\alpha}\mathop {\min }\limits_{i\in\{1,2\}}Q_{\boldsymbol{\phi}_i}\left(\mathbf{s}_{t},\cdot\right)\right) \right)\Bigg\}.
\end{align}
To further improve the sample efficiency and cope with policy collapse, a contrastive learning based auxiliary loss is leveraged to constrain the policy. As a result, the policy $\pi_{\boldsymbol{\theta}}$ with the parameters $\boldsymbol{\theta}$ for the actor network is optimized to minimize the augmented objective
\begin{align}\label{eq10}
\mathcal{L}_{\pi_{\boldsymbol{\theta}}}=&\mathbb{E}_{\mathbf{s}_{t} \sim \mathcal{D}, \mathbf{a}_t \sim \pi } \Big\{\alpha\log\left(\pi_{\boldsymbol{\theta}} \left( \left. \textbf{a}_{t} \right|\mathbf{s}_{t}\right)\right) - \mathop {\min }\limits_{i\in\{1,2\}}Q_{\boldsymbol{\phi}_i}\left(\mathbf{s}_{t},\mathbf{a}_t\right)\Big\}  \nonumber\\
&\quad\quad\quad\quad\quad\quad\quad\quad+\xi_4 \mathcal{L}_{\rm CL}\left(\mathbf{a}_t,\mathcal{A}_{\text{expert}}\right),
\end{align}
where the actions are sampled stochastically from  a squashed Gaussian policy given by $\mathbf{a}_t=\tanh\left(\mathbf{b}_{\boldsymbol{\theta}}\left(\mathbf{s}_{t}\right)+{\boldsymbol{\delta}}_{\boldsymbol{\theta}}\left(\mathbf{s}_{t}\right) \boldsymbol{\epsilon}\right)$ with the learnable bias $\mathbf{b}$, learnable scale matrix $\boldsymbol{\delta}$ and the input Gaussian noise vector $\boldsymbol{\epsilon} \sim \mathcal{N}\left(\mathbf{0},\mathbf{I}\right)$;  $\mathcal{L}_{\rm CL}$ is the contrastive loss with the weight $\xi_4$; $\mathcal{A}_{\text{expert}} = \{(\mathbf{s}_t^{\rm E}, \mathbf{a}_t^{\rm E})\}$ are the state-actions from the expert demonstrations. {As shown in Fig. 3, projection head is designed as a MLP (its parameters $\boldsymbol{\varphi}$) with two hidden layers containing 256 neurons at each hidden layer, to maximize the agreement between the representations of the expert behavior and the actor's policy via the contrastive loss.}  In the projection head, the actions $\mathbf{a}_t^{\rm E}$ from the expert and the actor's generated actions $\mathbf{a}\sim \pi \left( \left. \cdot\right|\mathbf{s}^{\rm E}_{t}\right)$ by observing the same states $\mathbf{s}^{\rm E}_{t}$ of the expert demonstrations are concatenated into the input samples, which are mapped to the latent space. Let $\mathbf{h}_{\rm E}$ denote the data points in the latent space corresponding to the expert's state-actions and $\mathbf{h}_{\rm A}$ denote the data points in the latent space corresponding to the actor's policy $\pi \left( \left. \mathbf{a}_t \right|\mathbf{s}^{\rm E}_{t}\right)$, the positive pair for a sample $j$ in a mini-batch is $\left(\mathbf{h}_{\rm E}\left(j\right),\mathbf{h}_{\rm A}\left(j\right)\right)$ under the same states $\mathbf{s}^{\rm E}_{t}$. Therefore, the contrastive loss is calculated as~\cite{chenting2020}
\begin{align}\label{eq11}
\mathcal{L}_{\rm CL}= - \frac{1}{B}\sum\limits_{j = 1}^B {\log \frac{{\exp \left( {\mathrm{sim}\left( \mathbf{h}_{\rm E}\left(j\right),\mathbf{h}_{\rm A}\left(j\right) \right)/\tau_{\rm CL}} \right)}}{{\sum\limits_{k = 1}^B {\exp \left( {\mathrm{sim}\left( \mathbf{h}_{\rm E}\left(k\right),\mathbf{h}_{\rm A}\left(k\right) \right)/\tau_{\rm CL}} \right)} }}},
\end{align}
where {$B$ is the batch size based on the data $\{\left(\mathbf{h}_{\rm E}\left(j\right),\mathbf{h}_{\rm A}\left(j\right)\right)\}$}; the cosine similarity is $\mathrm{sim}\left( \mathbf{h}_{\rm E},\mathbf{h}_{\rm A} \right)=\left\langle\mathbf{h}_{\rm E},\mathbf{h}_{\rm A}\right\rangle/\left(\left\| \mathbf{h}_{\rm E}\right\| \left\| \mathbf{h}_{\rm A}\right\|\right)$, and $\tau_{\rm CL}$ is the temperature parameter.

The actor network and the projection head are trained by jointly minimizing the loss $\mathcal{L}_{\pi_{\boldsymbol{\theta}}}$ and $\mathcal{L}_{\rm CL}$, when the training is completed, the projection head is thrown away. As such, the proposed algorithm is summarized in the \textbf{Algorithm 1}.

{\begin{algorithm}[t!]\footnotesize
\caption{RL learning algorithm with AR-based behavior cloning and contrastive learning}
\label{alg:my algo}
\begin{algorithmic}[1]
\Statex \textbf{Initialization}:  Expert demonstrations $\mathcal{A}_{\text{expert}} = \{(\mathbf{s}_t^{\rm E}, \mathbf{a}_t^{\rm E})\}$ via AR-based teleoperation, temperature parameters $\alpha$ and $\tau_{\text{CL}}$, hyperparameters, replay buffer, batch size,  the structures of the behavior cloning, target Q-network, critic, actor and projection head

\Statex \textbf{Phase 1: Behavior cloning pretraining}
\State The policy network $\pi_{\boldsymbol{\theta}}$ is pretrained by minimizing the objective  $\mathcal{L}_{\rm BC} = \mathbb{E} [\|\pi_{\theta}(\mathbf{a}_j|\mathbf{s}_j) - \mathbf{a}_j^{\rm E}\|^2]$, thus its parameters $\boldsymbol{\theta}$ are updated by $\boldsymbol{\theta} \leftarrow \boldsymbol{\theta} - \eta_{\rm BC} \nabla_{\boldsymbol{\theta}} \mathcal{L}_{\rm BC}$

\Statex \textbf{Phase 2: Soft Actor-Critic with projection head}
\For{each training iteration}
        \State  Sample  $\{(\mathbf{s}_{t},\mathbf{a}_t,r\left( {\mathbf{s}_t ,\mathbf{a}_t } \right),\mathbf{s}_{t+1}) \}\sim \mathcal{D}$
        \State  Sample $\{(\mathbf{s}_t^{\rm E}, \mathbf{a}_t^{\rm E})\}\sim \mathcal{A}_{\text{expert}}$
        \State The Q-network is trained by minimizing the Bellman residual $\mathcal{L}_{\boldsymbol{\phi}_i}$, thus its parameters are updated by  $\boldsymbol{\phi}_i\leftarrow \boldsymbol{\phi}_i - \eta_{Q_i} \nabla_{\boldsymbol{\phi}_i}    \mathcal{L}_{\boldsymbol{\phi}_i}$, $\forall i\in\{1,2\}$
        \State The projection head is trained by minimizing the contrastive loss $\mathcal{L}_{\rm CL}$,  thus its parameters $\boldsymbol{\varphi}$ are updated by $\boldsymbol{\varphi} \leftarrow \boldsymbol{\varphi} - \eta_{\boldsymbol{\varphi}} \nabla_{\boldsymbol{\varphi}} \mathcal{L}_{\rm CL}$
        \State After each gradient update of the projection head, the parameters of the actor network are updated by   $\boldsymbol{\theta} \leftarrow \boldsymbol{\theta} - \eta_{\pi} \nabla_{\boldsymbol{\theta}} \mathcal{L}_{\pi_{\boldsymbol{\theta}}}$
        \State Update the parameters of target Q-network $\boldsymbol{\bar{\phi}}_i \leftarrow \tau \boldsymbol{\phi}_i + (1 - \tau) \boldsymbol{\bar{\phi}}_i$, $\forall i\in\{1,2\}$
\EndFor
\State Remove the projection head, and \Return $\boldsymbol{\phi}_1$, $\boldsymbol{\phi}_2$, $\boldsymbol{\theta}$
\end{algorithmic}
\end{algorithm}}

\begin{figure*}
     \centering
    \subfigure[Physics simulations via Pybullet.]{
         \centering
         \includegraphics[width=3.5in,height=1.8in]{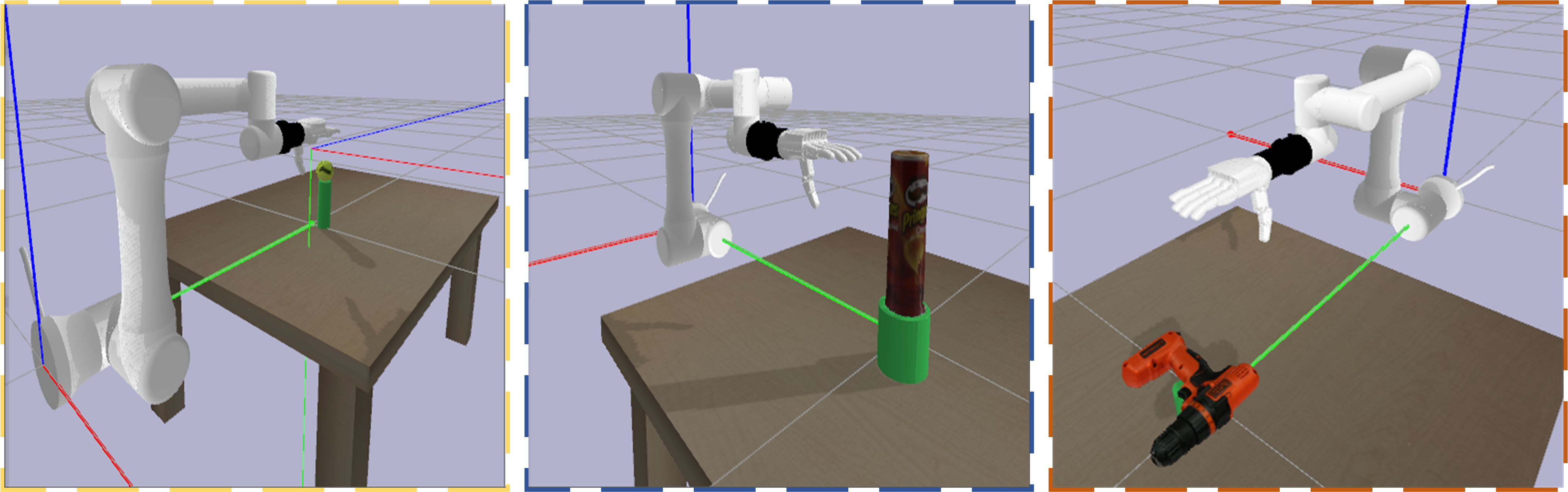}
      \label{fig4simulator}
     }
         \subfigure[Real-world demonstrations.]{
         \centering
         \includegraphics[width=2.0in,height=1.9in]{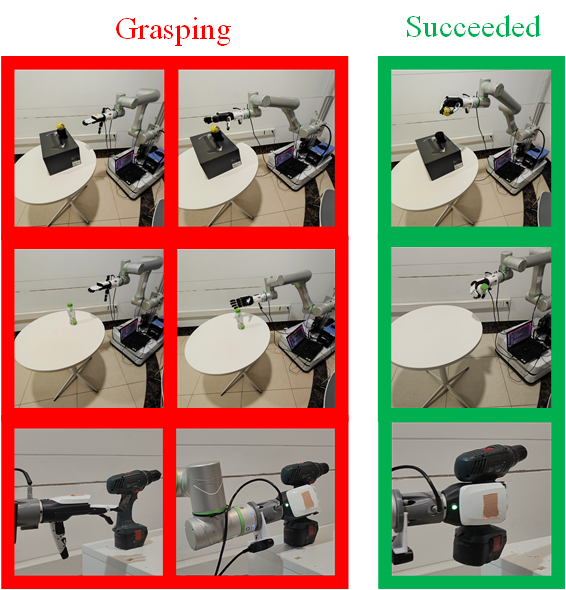}
      \label{fig4demo}
    }
 \caption{{Dexterous robot manipulation tasks in the simulation environment and real-world.}}
 \label{fig4}
\end{figure*}
\begin{figure*}
\centering
 \includegraphics[width=6.2in,height=2.0in]{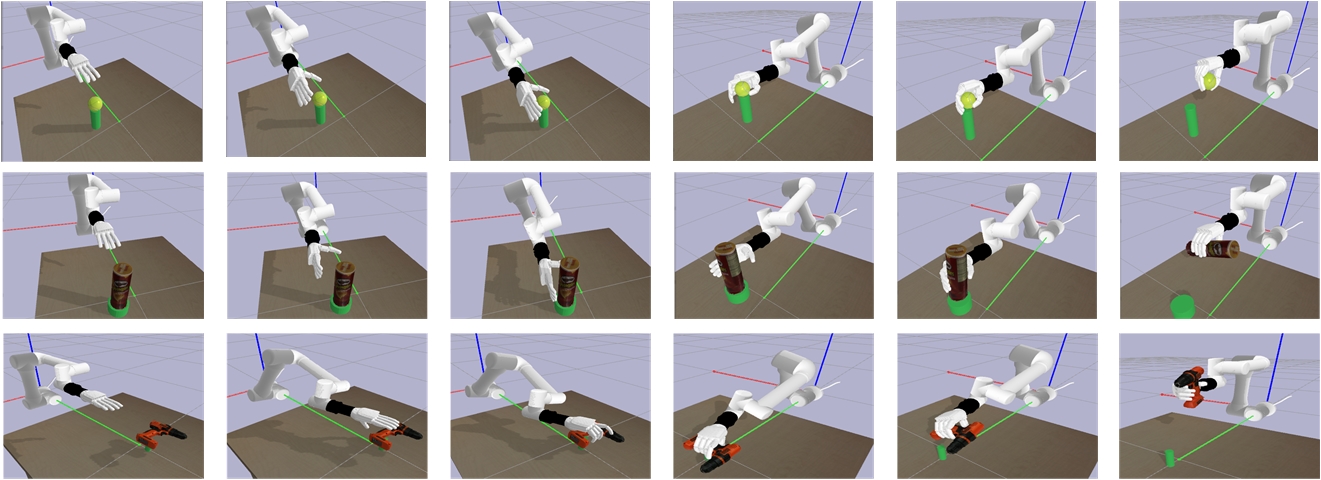}
\caption{{The rollouts of object grasping via the proposed algorithm: The functional grasps are achieved in the end.}}
 \label{fig5}
\end{figure*}

\section{Numerical Results}\label{sec:simulation}

This section provides numerical results to demonstrate the efficacy of the proposed algorithm through the PyBullet simulator in the Python environment and real-world trials as shown in Fig. 4 and
supplementary videos. In the considered dexterous robot arm-hand system with CGXi-G6 arm and an Inspire hand (See Fig. 1 and Fig. 7), we utilize the YOLOv8 object detection model for perception and conduct the object grasping tasks. During the network training, the Adam optimizer is adopted, the discount factor is $\gamma=0.98$, the learning rate is $7.3\times10^{-4}$, the batch size is $B=512$, and the parameters in the entropy augmented reward are outlined in Table I. The state $\mathbf{s}_t $ at each time step is a 20-element vector and the action $\mathbf{a}_t$ is an 8-element vector.  The parameters for the SAC  structures are outlined in Table II. The projection head is a MLP with two hidden layers (each layer has 256 hidden units) and an output feature size of 128, the temperature parameter of the contrastive loss is $\tau_{\rm CL}=0.1$, and the weight is $\xi_4=0.5$.
\begin{table}[]
\centering
\caption{Parameters in the entropy augmented reward}
\setlength\tabcolsep{2.5pt}
\scalebox{1.1}{\begin{tabular}{cc|cc}
\hline
 Parameters & Values & Parameters  & Values  \\ \hline
 $\alpha$ &  1 &  $\mathcal{Z}_1$ & 1000  \\ \hline
$\xi_1$ &  1000& $\mathcal{Z}_2$ & 100  \\ \hline
$\xi_2$ & 1 & $\mathcal{Z}_3$ & 60\\ \hline
$\xi_3$ &  1 & $\mathcal{Z}_4$ & 7\\ \hline
$\Lambda_{\rm th}$ & 0.75 & $\mathcal{Z}_5$ & 80 \\ \hline
\end{tabular} }
\end{table}
\begin{table}[]
\centering
\caption{Parameters for the SAC structure}
\setlength\tabcolsep{0.05pt}
\renewcommand{\arraystretch}{1.0}
\scalebox{1.1}{\begin{tabular}{c|c}
\hline
 Parameters & Values   \\ \hline
Number of hidden layers for all the networks &  2 \\ \hline
Number of hidden units per layer for all the networks & 256 \\ \hline
Replay buffer size & $3\times 10^5$ \\ \hline
Target smoothing coefficient & $\tau=0.005$  \\ \hline
\end{tabular} }
\end{table}

\begin{figure*}
\centering
 \includegraphics[width=6.6in,height=2.4in]{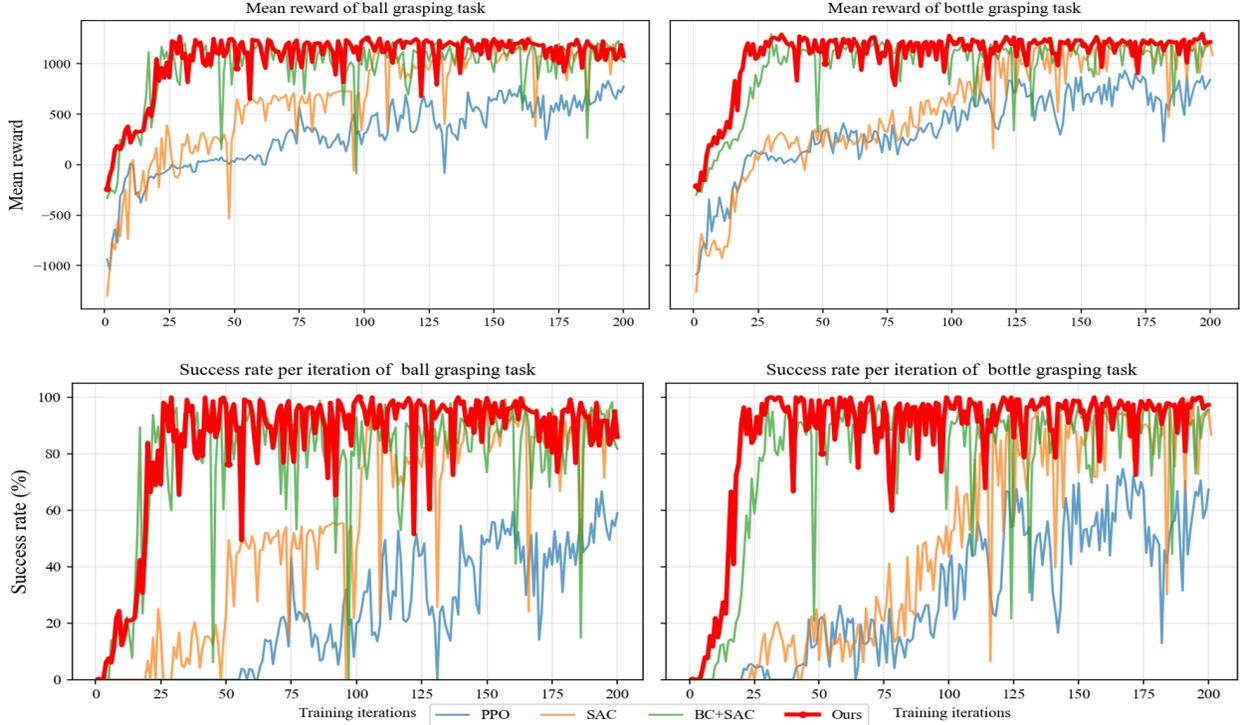}
\caption{{The convergence of using different RL algorithms for object grasping.}}
 \label{fig6}
\end{figure*}

{ In the simulations and real-world trials, three widely-used RL architectures are considered for performance comparison and the baselines are implemented as follows:
\begin{itemize}
  \item {PPO~\cite{PPO2017,Jianliang2026}: The policy is generated by a MLP with two hidden layers, and each hidden layer has 256 hidden units.}
  \item SAC~\cite{haarnoja18b}: The network structure stays the same as shown in Table II, which is also helpful for the ablation study.
  \item Advantage Weighted Actor-Critic (AWAC)~\cite{AshvinNair}:  Its actor is updated by maximizing the weighted likelihood policy, and the network structure also stays the same as shown in Table II.
  \item SAC with behavior cloning pretraining (BC$+$SAC): It is the case of our proposed algorithm without contrastive learning (absence of projection head component) for the sake of ablation study.
\end{itemize} }

\begin{table*}[htbp]
    \centering
    \setlength\tabcolsep{2.5pt}
    \caption{{ BC performance for different numbers of expert trajectories (Traj.)}}
    \label{tab:bc_ablation}
    \begin{tabular}{|l|l|c|c|c|c|c|}
        \hline
        \textbf{Tasks} & \multicolumn{1}{|c|}{\textbf{Metrics}} & \textbf{5 Traj.} & \textbf{10 Traj.} & \textbf{15 Traj.} & \textbf{30 Traj.} & \textbf{50 Traj.} \\ \hline

        \multirow{2}{*}{Bottle grasping}
        & Mean reward and standard deviation& 110.12$\pm$75.4 & 285.30$\pm$45.1 & 440.23$\pm$31.45 & 692.45$\pm$25.3 & 690.12$\pm$18.8 \\ \cline{2-7}
        & Mean success rate and standard deviation& 8.20\%$\pm$4.5\% & 24.50\%$\pm$3.2\% & 40.15\%$\pm$1.8\% & 53.45\%$\pm$2.5\% & 53.12\%$\pm$2.1\% \\ \hline

        \multirow{2}{*}{Ball grasping}
        & Mean reward and standard deviation& 95.40$\pm$82.8 & 240.15$\pm$55.5 & 372.45$\pm$38.62 & 548.30$\pm$32.1 & 565.45$\pm$24.5 \\ \cline{2-7}
        & Mean success rate and standard deviation& 6.45\%$\pm$3.2\% & 20.12\%$\pm$3.5\% & 33.47\%$\pm$2.1\% & 42.60\%$\pm$3.8\% & 43.47\%$\pm$3.4\% \\ \hline

        \multirow{2}{*}{Drill grasping}
        & Mean reward and standard deviation& 52.15$\pm$95.2 & 145.30$\pm$60.8 & 229.12$\pm$45.30 & 415.60$\pm$40.2 & 440.10$\pm$32.6 \\ \cline{2-7}
        & Mean success rate & 3.50\%$\pm$2.8\% & 12.30\%$\pm$4.9\% & 21.60\%$\pm$3.2\% & 32.94\%$\pm$2.7\% & 34.35\%$\pm$3.1\% \\ \hline
    \end{tabular}
\end{table*}

\begin{table*}[]
\centering
\caption{{Performance comparisons under different manipulation tasks and metrics}}
\setlength\tabcolsep{1.2pt}
\begin{tabular}{|c|l|c|c|c|c|c|c|}
\hline
Tasks & \multicolumn{1}{c|}{Metrics} & {BC}  & PPO    & SAC & BC+SAC & {AWAC} &\textbf{Ours} \\ \hline
\multirow{4}{*}{Bottle grasping}& {Mean reward and standard deviation}                             &440.23$\pm$31.45  &969.20$\pm$58.72 & 1036.50$\pm$112.74&1129.52$\pm$44.33& 1182.21$\pm$55.43 &1251.94$\pm$65.62 \\ \cline{2-8}
                                 & {Mean success rate  and standard deviation }                     &40.15\%$\pm$1.8\% & 79.30\%$\pm$2.5\% & 84.85\%$\pm$4.6\% &89.00\%$\pm$2.4\% & 93.53\%$\pm$2.5\% & 95.70\%$\pm$2.8\% \\ \cline{2-8}
                                 & Number of iterations until convergence   &  N/A      &  210   &   130     &  30 & 35& 30   \\ \cline{2-8}
                                 & Time consumption until convergence (min) &   2     &   500  &    280   &   75 & 85  & 75\\ \hline
\multirow{4}{*}{Ball grasping}& {Mean reward and standard deviation}                          & 372.45$\pm$38.62 & 860.77$\pm$66.92 & 1021.49$\pm$88.69 &  1020.87$\pm$94.54& 1105.15$\pm$71.23 &  1195.40$\pm$65.73 \\ \cline{2-8}
                                 & {Mean success rate  and standard deviation  }                    &  33.47\%$\pm$2.1\%  & 69.27\%$\pm$2.6\%   & 83.46\%$\pm$3.0\%     &84.36\%$\pm$3.5\%   & 88.36\%$\pm$3.2\%& 91.39\%$\pm$2.9\% \\ \cline{2-8}
                                 & Number of iterations until convergence   &  N/A     & 220   &   150    &   60 & 60 & 50 \\ \cline{2-8}
                                 & Time consumption until convergence (min) &   2     &  550   &   420    &  150 & 150& 100  \\ \hline
\multirow{4}{*}{Drill grasping}& {Mean reward  and standard deviation }                           & 229.12$\pm$45.30 & 808.03$\pm$57.02 & 893.26$\pm$111.07& 909.97$\pm$135.85 &955.45$\pm$48.57 & 1023.11$\pm$67.35 \\ \cline{2-8}
                                 & {Mean success rate and standard deviation }                      &  21.60\%$\pm$3.2\% & 64.90\%$\pm$4.0\%  & 76.65\%$\pm$6.8\% &77.15\%$\pm$5.3\% &81.26\%$\pm$4.5\% & 85.30\%$\pm$4.2\%\\ \cline{2-8}
                                 & Number of iterations until convergence   &  N/A     & 260   &   200    &   180& 90 & 70 \\ \cline{2-8}
                                 & Time consumption until convergence (min) &   2     &  650   &   500    &  450 & 200 & 170  \\ \hline
\end{tabular}
\end{table*}

\subsection{Manipulation Tasks for Object Grasping}

{We consider three cases, namely bottle, deformable ball and drill grasping tasks.  Both the proposed algorithm and the baselines are trained in the simulator via PyBullet (See Fig.~\ref{fig4simulator}), and the maximum trajectory length is up to $T=100$ time-steps. In each simulated trial, we run 3 seeds with seed number 42, 97, 215, and quantifies the performance for all the solutions  over $10^4$ simulated trials. Moreover, we have considered that the objects are positioned in dynamic 3D configurations, to exhibit spatial generalization. Fig. 5 presents the example rollouts for successful object grasping tasks. }

{By leveraging our dexterous robot arm-hand teleoperation system via AR in the subsection II-B, expert trajectories are collected for BC training. As shown in Table III, adding expert trajectories improves BC performance, but BC still performs poorly even with 50 expert trajectories. To strike a balance between BC performance and the cost of collecting demonstrations, we adopt 15 expert trajectories for BC training/pretraining in the following results.}

 Fig. 6 confirms that in all the use cases, our proposed algorithm converges more rapidly than the baselines, and obtains the largest mean reward, therefore, it achieves the highest success rates for object grasping tasks. With the help of behavior cloning based pretraining, the proposed algorithm significantly improves the reward and success rate from a few data points. The RL policy collapse can be well prevented by applying the contrastive learning in our proposed algorithm.

We proceed to evaluate the average performance in the simulation environment via PyBullet, where all the simulations are performed on the Intel Xeon Gold 6226R CPU and a single NVIDIA RTX 4090 GPU. {In each simulated trial, we run 3 seeds with seed number 42, 97, 215, and quantifies the performance for all the solutions  over $10^4$ simulated trials in terms of different metrics. The comprehensive performance comparisons are summarized in  Table IV. It is demonstrated that our proposed algorithm consumes the least training time\footnote{{Note that the time of expert demonstration data collection via AR based teleoperation is excluded, and the training duration of the proposed algorithm includes both BC pretraining and RL phases.}}  and achieves the highest success rates for all the considered manipulation tasks.}

We implement the proposed algorithm and baselines in a practical dexterous robot arm-hand system for real-world trials, moreover, an AR-based visual interface is created to remotely start the dexterous robot (See the supplementary videos) for manipulation tasks as shown Fig. 7. {In Fig. 8, the success rate for object grasping is evaluated over 50 real-world trials. The proposed algorithm achieves the highest success rates of bottle, deformable ball and drill grasping tasks, compared to the baselines.} Compared to the results obtained from the physics simulations in Table IV, the success rates of deformable ball (sphere) and drill grasping decrease in the real-world trials, due to the imperfect perception  and object's complex geometric features~\cite{ZhaoNMI} (Note that during the training via the simulator, the objects are very coarsely formed in the physics simulations without high-fidelity object geometry reconstruction and analyzing elastomer surface deformation~\cite{ZhaoNMI}). Since the bottles (cylinders) are easier to grasp than the deformable spheres in practice, the achievable success rate of bottle grasping for the proposed algorithm is still promising in the real-world trials, namely marginal sim-to-real gap is achieved.
\begin{figure}
\centering
 \includegraphics[width=3.2in,height=2.2in]{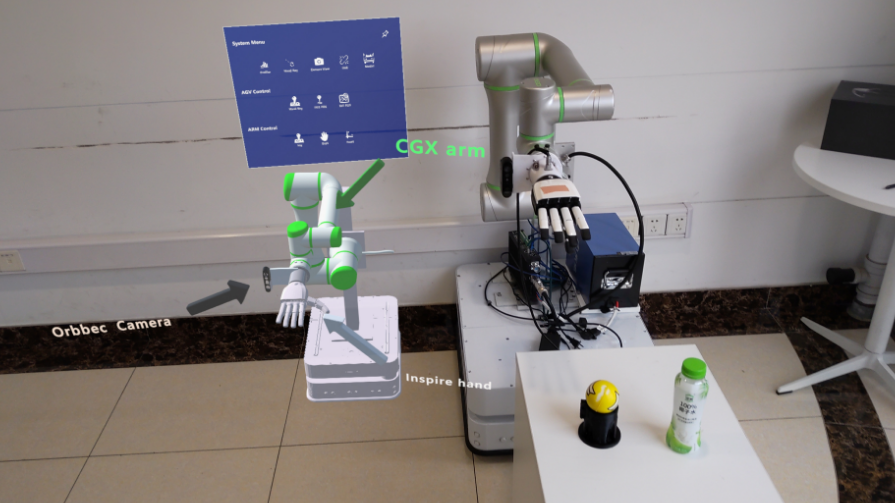}
\caption{AR-based visual interface overlaid onto the real-world.}
 \label{fig7}
\end{figure}

\begin{figure}
\hspace{-0.3cm}
\centering
 \includegraphics[width=3.6in,height=2.6in]{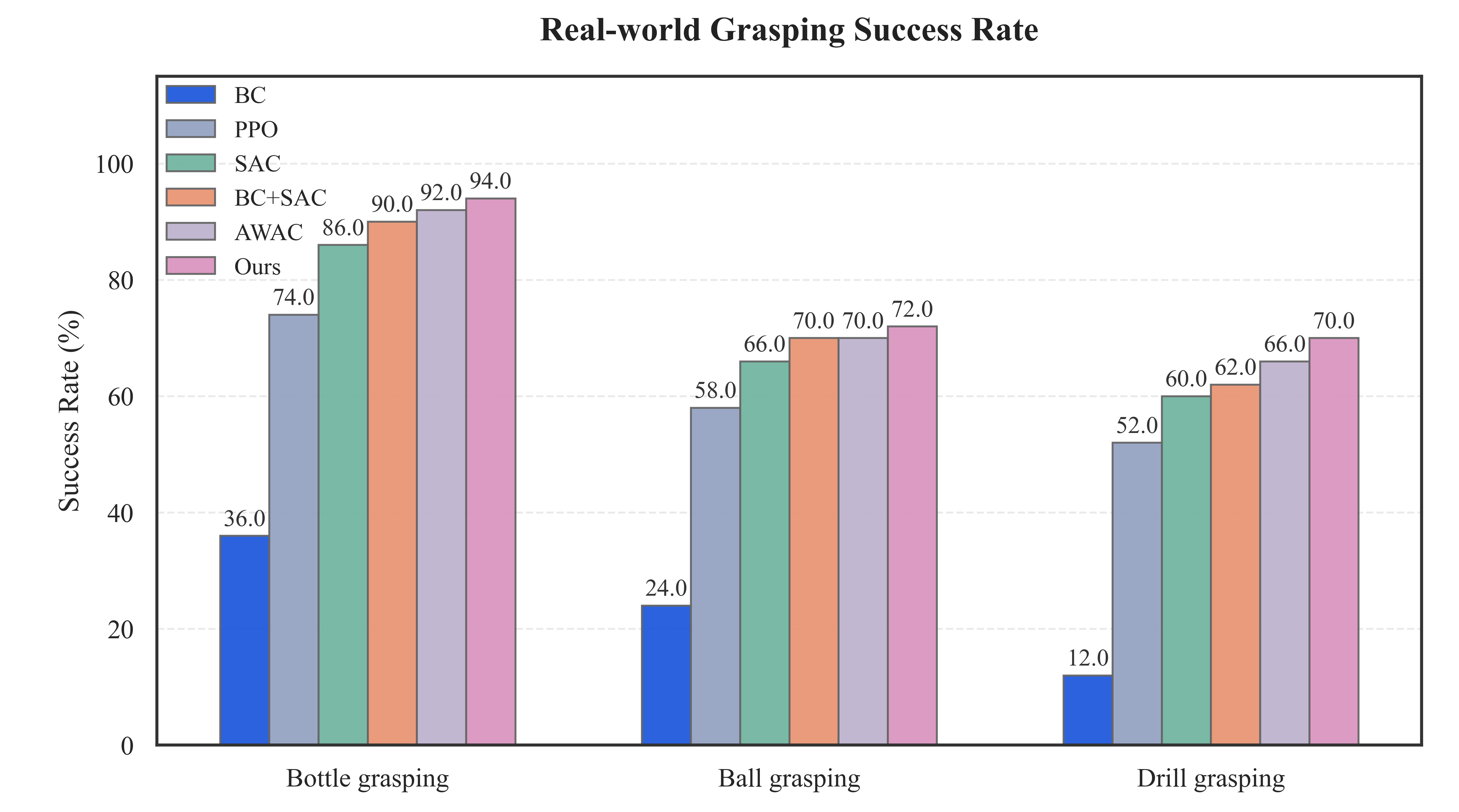}
\caption{{The performance comparisons in the real-world trials.}}
 \label{fig8}
\end{figure}

\section{Conclusions and Future Work}\label{conclusion_section}
In the dexterous robot arm-hand systems, manipulation task problem was addressed. A novel model-free RL algorithm was proposed with the help of imitation and contrastive learning and the RL training went through two phases. In the pretraining phase, a behavior cloning based policy was generated by solving a simple regression problem. To collect the human demonstration data, we constructed an AR-based teleoperated dexterous robotic system, which can support remote human-robot interactions. After the policy pretraining, the RL policy was improved by combining the SAC with contrastive learning, in particular, a projection head was introduced to improve the similarity between the agent's actions and expert's actions, which could overcome the policy collapse. The results showed that the proposed model-free algorithm rapidly reached convergence and avoided policy collapse compared to the baselines including PPO, SAC and AWAC. Both the physics simulation results and real-world trials demonstrated that the proposed algorithm obtained the highest success rate of completing the  manipulation task compared to the baselines.

In the future work, { training policies jointly across different physics simulators helps enhance the system robustness~\cite{Lei2025PolySimBT}. The sim-to-real gap could be  narrowed by considering high-fidelity object geometry reconstruction in the physics simulation and improving the perception.  Recent works such as~\cite{tang2025visualgeometry} fused the 2D semantics with 3D geometry for spatial feature alignment, which can be adopted to further improve the accuracy of robot learning.}  Moreover, the visual and haptic feedback  may need to be employed for accomplishing complicated deformable object manipulation. In the dexterous dual arm-hand robotic systems, the line of our work could be extended to the multi-agent case where the interactions between agents are required. {Residual RL is another important research direction, in which it is necessary to design an efficient BC network architecture as the base policy~\cite{LAnkile2024}}. {Furthermore, in unstructured real-world environments,  deploying RL alongside multimodal imitation learning remains an open research area.}


\bibliographystyle{IEEEtran}

\end{document}